\documentclass{article}
\pdfoutput=1

\PassOptionsToPackage{numbers, compress}{natbib}

\usepackage{multirow}
\usepackage{adjustbox}
\usepackage{subcaption}
\usepackage{enumitem}

\usepackage[preprint]{style}

\usepackage[utf8]{inputenc} 
\usepackage[T1]{fontenc}    
\usepackage[colorlinks,linkcolor=red,citecolor=blue,urlcolor=magenta,bookmarks=true]{hyperref}       
\usepackage{url}            
\usepackage{booktabs}       
\usepackage{amsfonts}       
\usepackage{nicefrac}       
\usepackage{microtype}      
\usepackage{xcolor}         
\usepackage{subcaption}
\usepackage{comment}
\usepackage{floatrow}
\newfloatcommand{capbtabbox}{table}[][\FBwidth]
\usepackage{array}
\usepackage{multirow}
\usepackage{wrapfig}
\usepackage{bm}
\usepackage{amsmath}
\usepackage{color}
\usepackage{diagbox}
\usepackage{amssymb}
\usepackage{setspace}
\usepackage{pifont}
\usepackage{adjustbox}
\usepackage{graphicx}

\title{Playground v2.5:  Three Insights towards Enhancing Aesthetic Quality in Text-to-Image Generation}

\author{%
  Daiqing Li \quad
  \quad Aleks Kamko 
  \quad Ehsan Akhgari
  \quad Ali Sabet 
  \quad Linmiao Xu
  \quad Suhail Doshi
  \vspace{4mm}\\
  Playground Research
  \vspace{4mm}
}

\begin{document}

\maketitle

\maketitle
\begin{center}
\vspace{-9mm}
    \centering
  	\captionsetup{type=figure}
	\includegraphics[width=1.0\textwidth]{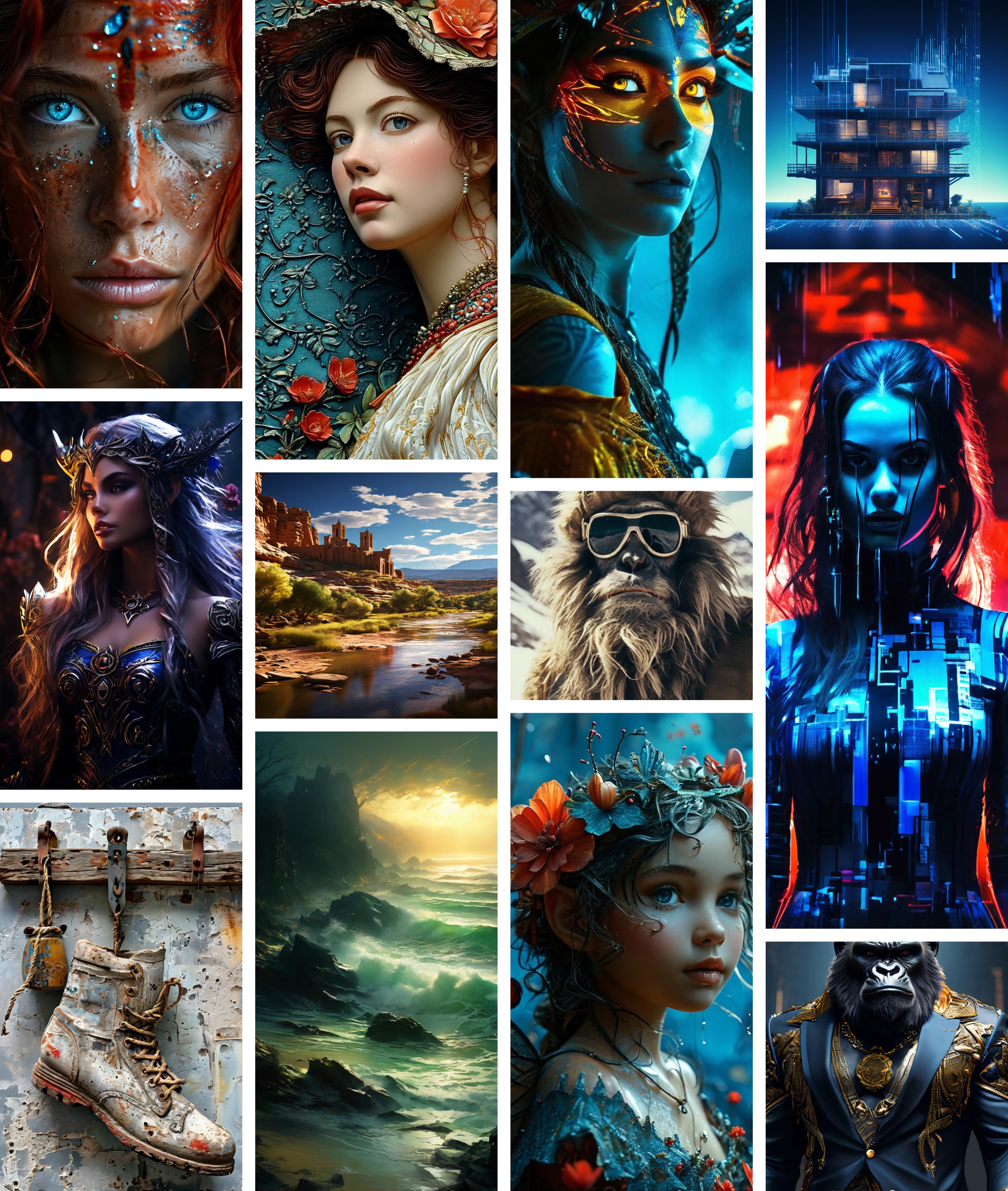}
 
	\captionof{figure}{\small {\textbf{High-quality samples from Playground v2.5.} Our model exhibits vibrant color and contrast on a range of image styles. Samples generated by the Playground community members.}}
	\label{fig:teaser}

\end{center}

\newpage

\begin{abstract}
In this work, we share three insights for achieving state-of-the-art aesthetic quality in text-to-image generative models. 
We focus on three critical aspects for model improvement: enhancing color and contrast, improving generation across multiple aspect ratios, and improving human-centric fine details. First, we delve into the significance of the noise schedule in training a diffusion model, demonstrating its profound impact on realism and visual fidelity.
Second, we address the challenge of accommodating various aspect ratios in image generation, emphasizing the importance of preparing a balanced bucketed dataset. Lastly, we investigate the crucial role of aligning model outputs with human preferences, ensuring that generated images resonate with human perceptual expectations. Through extensive analysis and experiments, Playground v2.5 demonstrates state-of-the-art performance in terms of aesthetic quality under various conditions and aspect ratios, outperforming both widely-used open-source models like SDXL \cite{podell2023sdxl} and Playground v2 \cite{playground-v2}, and closed-source commercial systems such as DALL{$\cdot$}E 3 \cite{betker2023improving} and Midjourney v5.2. Our model is open-source, and we hope the development of Playground v2.5 provides valuable guidelines for researchers aiming to elevate the aesthetic quality of diffusion-based image generation models.
\end{abstract}
\begin{figure*}[th!]
\vspace{-5mm}
\centering
\begin{subfigure}[b]{1.0\textwidth}
\includegraphics[width=1.0\textwidth]{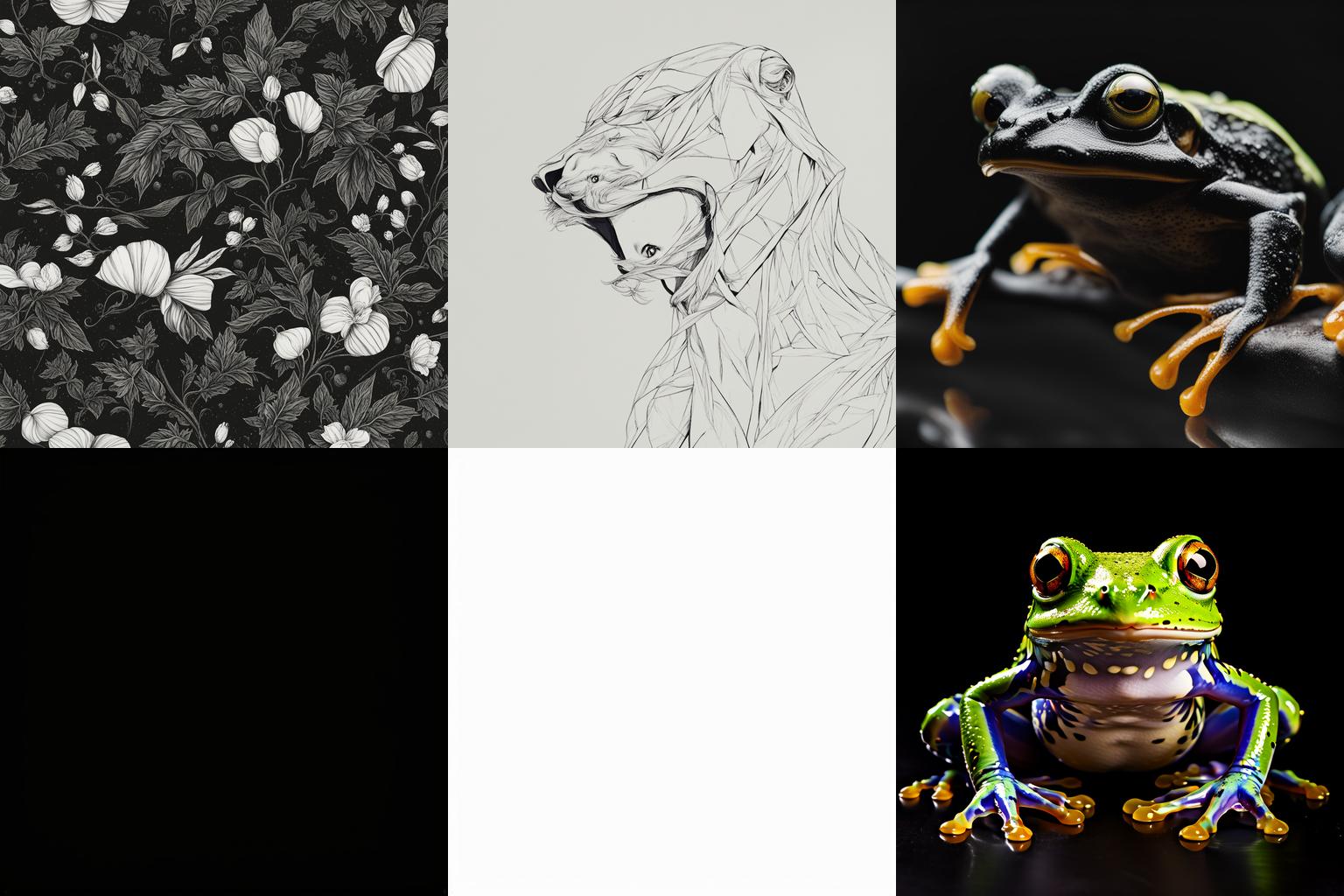}
\vspace{2mm}
{\setstretch{0.7}
\begin{tabular}{p{40mm}p{40mm}p{40mm}}
    {\scriptsize{a solid black background}} &
    {\scriptsize{a solid white background}} &  
    {\scriptsize{frog in front of solid black background}} 
\end{tabular}
}

\caption{\textbf{Generating solid backgrounds.}
The top row is sampled from SDXL~\cite{podell2023sdxl}, bottom row is Playground v2.5. SDXL fails to generate pure black or white background while our model can follow the prompt faithfully.}
\end{subfigure}
\begin{subfigure}[b]{1.0\textwidth}
\includegraphics[width=1.0\textwidth]{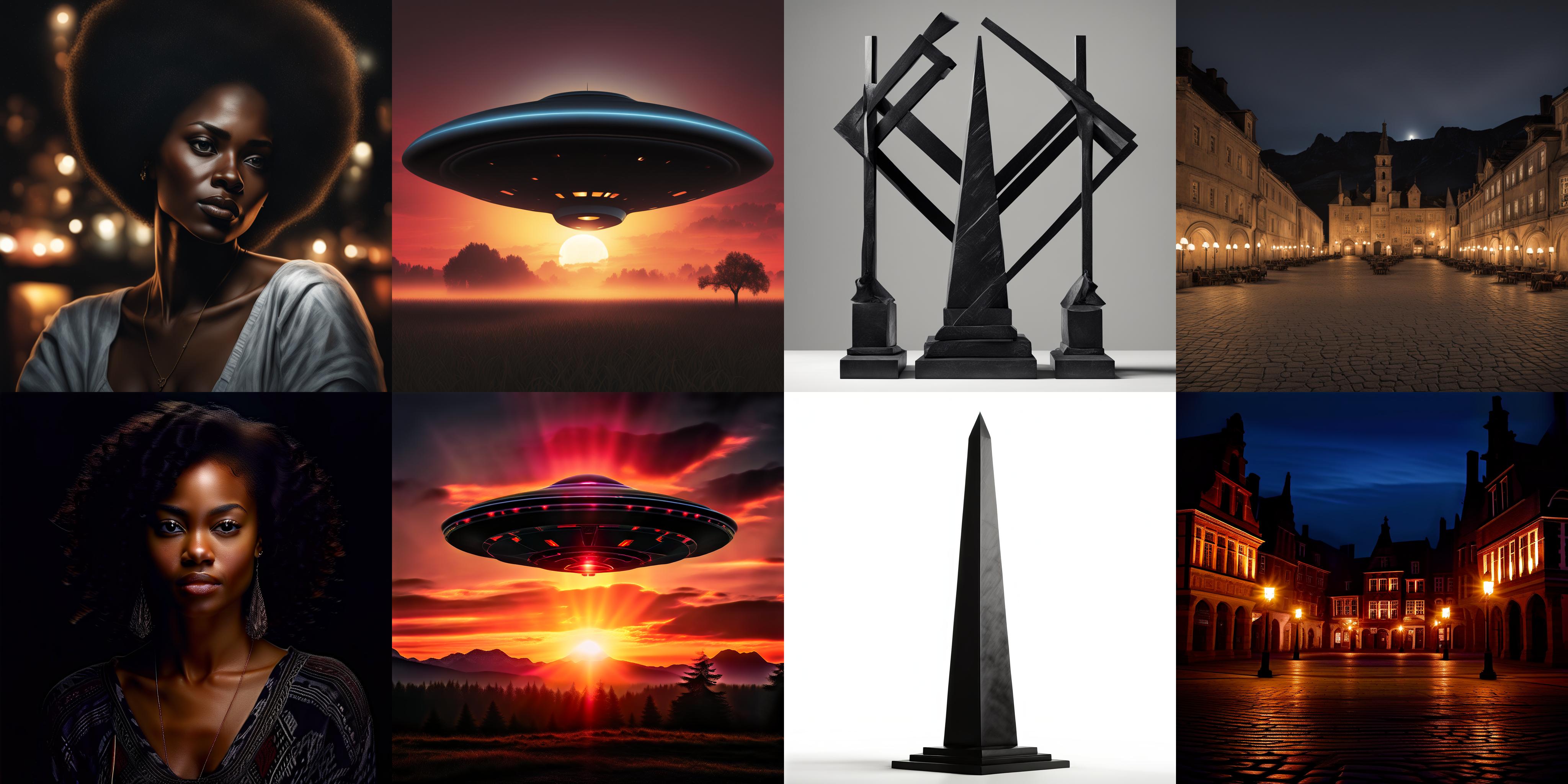}
\vspace{2mm}
{\setstretch{0.7}
\begin{tabular}{p{30mm}p{30mm}p{30mm}p{30mm}}
    {\scriptsize{A portrait of an African American woman on a black background at night, photo realistic, ultra realistic}} &
    {\scriptsize{Black UFO in front of a rising sun at the dawn, vibrant colours, dark surroundings, photorealisti}} &  
    {\scriptsize{A black obelisk in front of a white background}}  &
    {\scriptsize{A town square lit only by torchlight}} 
\end{tabular}
}

\caption{\textbf{Colors and contrast}. The top row is SDXL, bottom row is Playground v2.5. Our model can generate samples with more vibrant colors and contrast.
}
\end{subfigure}
\caption{\textbf{Comparing SDXL and Playground v2.5 in generating images with vibrant color and contrast.}
}
\label{fig:color_contrast}
\end{figure*}
\vspace{-4mm}
\section{Introduction}

Great progress has been made in diffusion-based generative models since the success of better image modeling performance \cite{ho2020denoising,song2021scorebased,dhariwal2021diffusion}
with ImageNet, as compared to performance with the previously dominate framework of generative adversarial networks (GAN) \cite{goodfellow2014generative, karras2019stylebased, karras2020analyzing}. Open-source models like SDXL~\cite{podell2023sdxl} have built on top of latent diffusion models (LDM)~\cite{rombach2022highresolution} by scaling up text-to-image pre-training datasets~\cite{schuhmann2022laion5b} and the latent UNet~\cite{dhariwal2021diffusion} architecture. PixArt-alpha~\cite{chen2023pixart}, on the other hand, explores Diffusion Transformer (DiT) \cite{peebles2023scalable} as the latent backbone, showing better training efficiency and image quality. Playground v2 \cite{playground-v2}, an open-source model recently developed by us, focuses on the training recipe and aesthetic quality, achieving $2.5\times$ higher user preference compared to SDXL~\cite{podell2023sdxl}.  

Playground v2~\cite{playground-v2} was open-sourced in December 2023 and we were pleased to see the open-source and research community take up our work and reference it. Notably, Playground v2 has amassed over 135,000 downloads in just the last month from HuggingFace,
and our work has been cited in recent papers for state-of-the-art image models such as Stable Cascade \cite{stablecascade}. 
Following Playground v2 \cite{playground-v2}, we chose not to change the underlying model architecture for this work; instead, we focused on analyzing and improving our training recipe and pushing the model's aesthetic quality to a new level. 

We focus on three critical issues for image models: enhancing color and contrast (sec.~\ref{color}), improving generation across multiple aspect ratios (sec.~\ref{multiar}), and improving human-centric fine details (sec.~\ref{human_pref_alignment}). More generally, we aim to refine the model’s capabilities to produce more realistic and visually compelling outputs. To evaluate the efficacy of our enhancements, we conducted extensive user studies and benchmarked our model against previous state-of-the-art models (sec.~\ref{eval}). We also propose a new automatic-evaluation benchmark \textit{MJHQ-30K} (sec.~\ref{mjhq30k}) to evaluate the model's performance against 10 unique categories. 
When evaluating our model’s outputs on human preference, we are thrilled to report that Playground v2.5 surpasses state-of-the-art models, including Midjourney 5.2, DALL{$\cdot$}E 3~\cite{betker2023improving}, Playground v2~\cite{playground-v2}, PIXART-$\alpha$~\cite{chen2023pixart}, and SDXL~\cite{podell2023sdxl} (Fig \ref{fig:sota}). See sec. ~\ref{sota_study} for details. Playground v2.5 endeavors to surpass the performance of its predecessor and establish itself as a leading contender in the space of text-to-image generative models.

We open-source the weights for Playground v2.5, on HuggingFace\footnote{\url{https://huggingface.co/playgroundai/playground-v2.5-1024px-aesthetic}} with a license\footnote{\url{https://huggingface.co/playgroundai/playground-v2.5-1024px-aesthetic/blob/main/LICENSE.md}} that makes it easy for research teams to use. We will also provide extensions for using our model in A1111 and ComfyUI, two popular community tools for using diffusion models. Given how much we have benefited from the research and open-source communities, it is important that we make multiple aspects of our work on Playground v2.5 available to the community. 

\begin{figure*}[!t]
\centering

\includegraphics[width=1.0\textwidth]{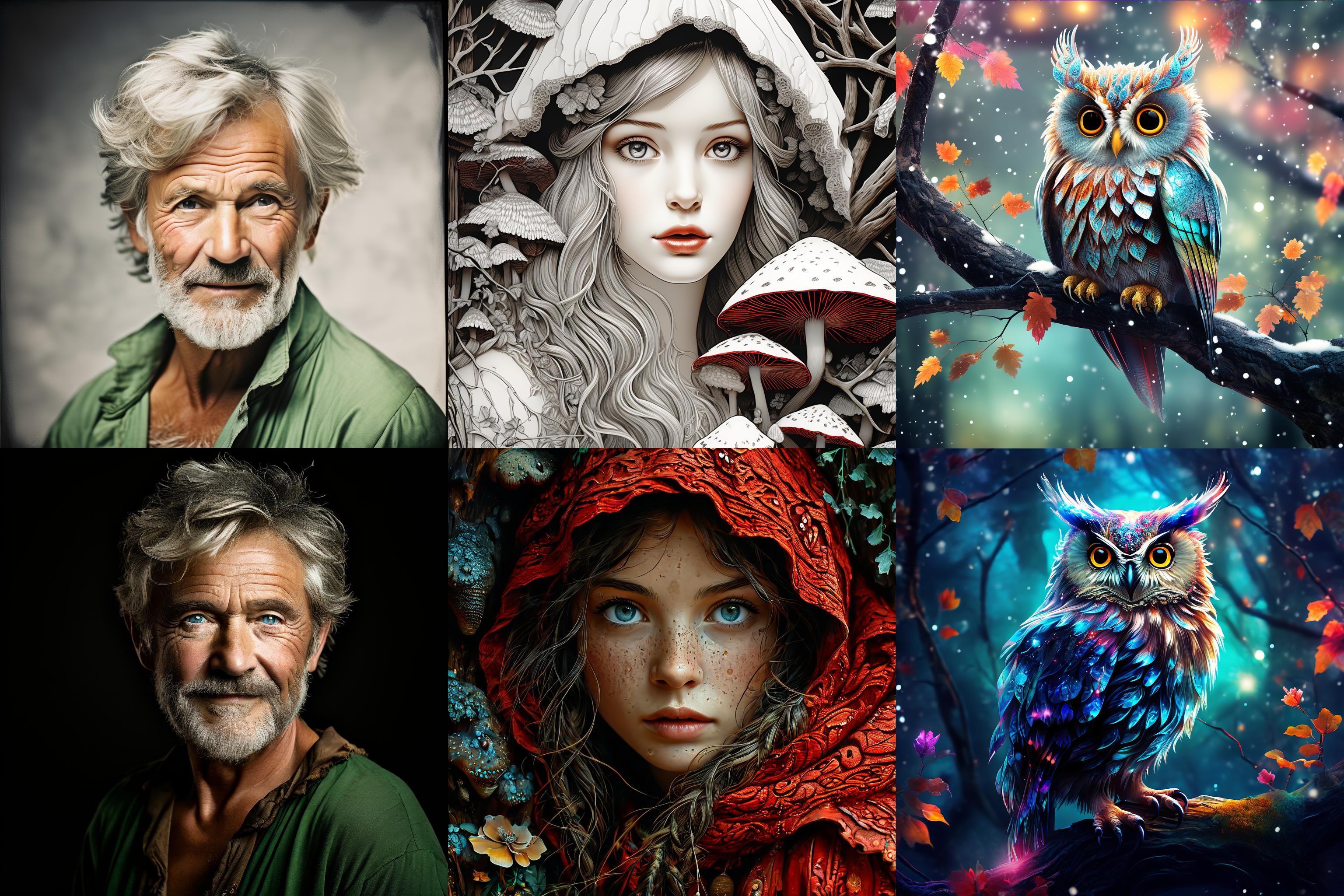}

\vspace{2mm}
{\setstretch{0.7}
\begin{tabular}{p{40mm}p{40mm}p{40mm}}
    {\scriptsize{Peter Pan aged 60 years old, with a black background}} &
    {\scriptsize{Bilibin ink detailed masterpiece, National Geographic gracious, filigree acrylic, Slavic folklore, tender face, storybook illustration, art on a cracked wood, young beautiful Red Riding Hood girl, book illustration style, forest, mushrooms, hyperrealism, digital art, cinematic, close portrait, highly detailed expressive glowing eyes, airy, detailed face, shadow play, realistic textures, dynamic pose, unusual, modern. heartwarming, cozy, fairytale, fantasy, detailed textures, artistic dynamic pose, tender, atmospheric, sharp focus, centered composition, complex background, soft haze, masterpiece. animalistic, beautiful, tiny detailed}} &  
    {\scriptsize{happy dreamy owl monster sitting on a tree branch, colorful glittering particles, forest background, contoured, surrealism, close up cute, detailed  feathers, bioluminescence, leaves, ethereal, ice, looking in camera, sky, sleek, modern, fairytale, fantasy, by Andy Kehoe}} 
\end{tabular}
}

\caption{\textbf{Comparing Playground v2~\cite{playground-v2} and v2.5 for color and contrast with more complex prompts.} Top row is Playground v2, bottom row is Playground v2.5. Compared to v2, v2.5 dramatically improves the color and contrast, and the ability to follow style-related prompts.
}

\label{fig:color-peter}
\end{figure*}
\vspace{-4mm}
\section{Methods}
\subsection{Enhanced Color and Contrast}
\label{color}
Latent diffusion models have struggled to generate images with high color contrast and vibrant color range since the release of SD1.5. This is a known limitation \cite{lin2024common, chen2023importance, hoogeboom2023simple}. For example, SDXL is incapable of generating a pure black image or a pure white image, and fails to place subjects onto solid backgrounds (see Fig~\ref{fig:color_contrast} (a)).

This issue stems from the noise scheduling of the diffusion process, as pointed out by \cite{lin2024common}. The signal-to-noise ratio \cite{kingma2023variational} of Stable Diffusion is too high, even when the discrete noise level reaches its maximum. Several works have attempted to fix this flaw. Guttenberg and CrossLabs \cite{offsetnoise} propose offset noise. Lin et al.~\cite{lin2024common} propose Zero Terminal SNR to ensure the last denoising step is pure Gaussian noise. SDXL \cite{podell2023sdxl} adopts the strategy of adding offset noise in the last stage of the training, as does Playground v2. However, as can be seen in Fig~\ref{fig:color_contrast} (b), we still notice SDXL exhibits muted color and contrast.
\begin{figure*}
\includegraphics[width=1.0\textwidth]{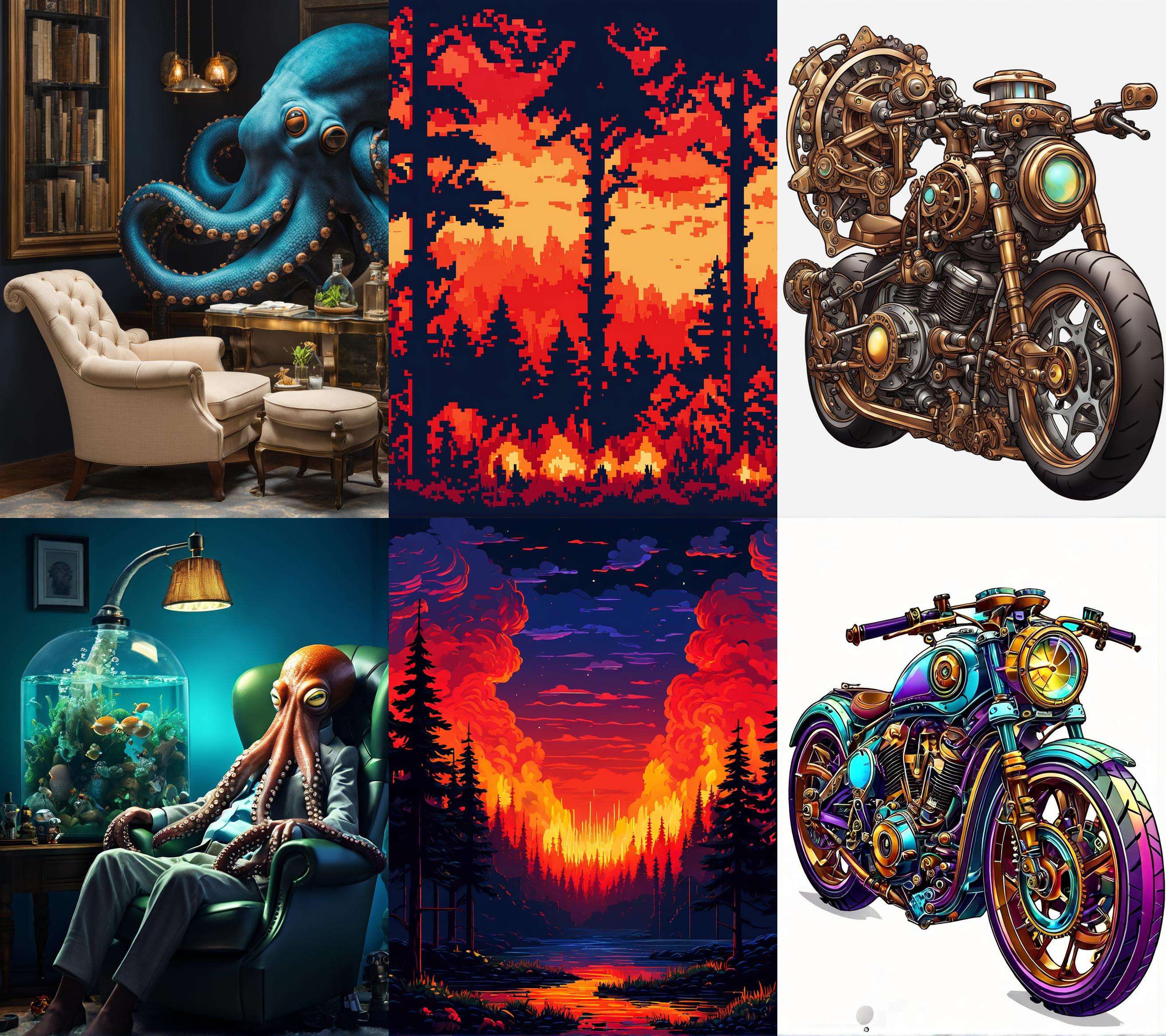}
\begin{tabular}{ccc}
    \multirow{2}{40mm}{\scriptsize{human-like octopus sitting in a recliner with a human in fish tank on his side table.}} &  \multirow{2}{40mm}{\scriptsize{8-bit color forest
fire visual effect}} &  \multirow{6}{40mm}{\scriptsize{teampunk vivid illuminated iridescent intricate mechanical racing motorcycle
with intricate brown cogwheels sticker, white background, contour, colorful, vector, kawaii, hdr,
Watercolor, trending on artstation, sharp focus, studio photo, intricate details, highly detailed, by greg
rutkowski"}} 
\end{tabular}
\vspace{22mm}

\caption{\textbf{Qualitative comparison of portrait aspect ratios.} Aspect ratio 3:4 ($876\times1168$), top is SDXL, bottom is Playground v2.5. Our model can generate images with the correct composition to the desired aspect ratio 
}
\label{fig:multiar-oct}
\end{figure*}

For Playground v2.5, we aimed to dramatically improve upon this issue. We wanted to achieve vivid color and contrast in imagery and be able to produce pure-colored backgrounds. To this end, we took a more principled approach and trained our models from scratch using the EDM framework, proposed by Karras et al \cite{karras2022elucidating}.

EDM brings two distinct advantages: (1) Like Zero Terminal SNR, the EDM noise schedule exhibits a near-zero signal-to-noise ratio for the final “timestep”. This removes the need for Offset Noise and fixes muted colors. (2) EDM takes a first-principles approach to designing the training and sampling processes, as well as preconditioning of the UNet. This enables the EDM authors to make clear design choices that lead to better image quality and faster model convergence.

We were also inspired by Hoogeboom et al \cite{hoogeboom2023simple} to skew the noise schedule towards an overall noisier one when training on high-resolution images.

In Fig \ref{fig:color-peter}, we show a qualitative comparison between Playground v2.5 and Playground v2, the latter of which uses offset noise and a DDPM \cite{ho2020denoising} noise schedule. We see in the first column that Playground v2.5 can generate a vivid portrait with enhanced color range, and it exhibits better prompt-image alignment, which enables v2.5 to generate a pure black background.

\begin{figure*}
\includegraphics[width=1.0\textwidth]{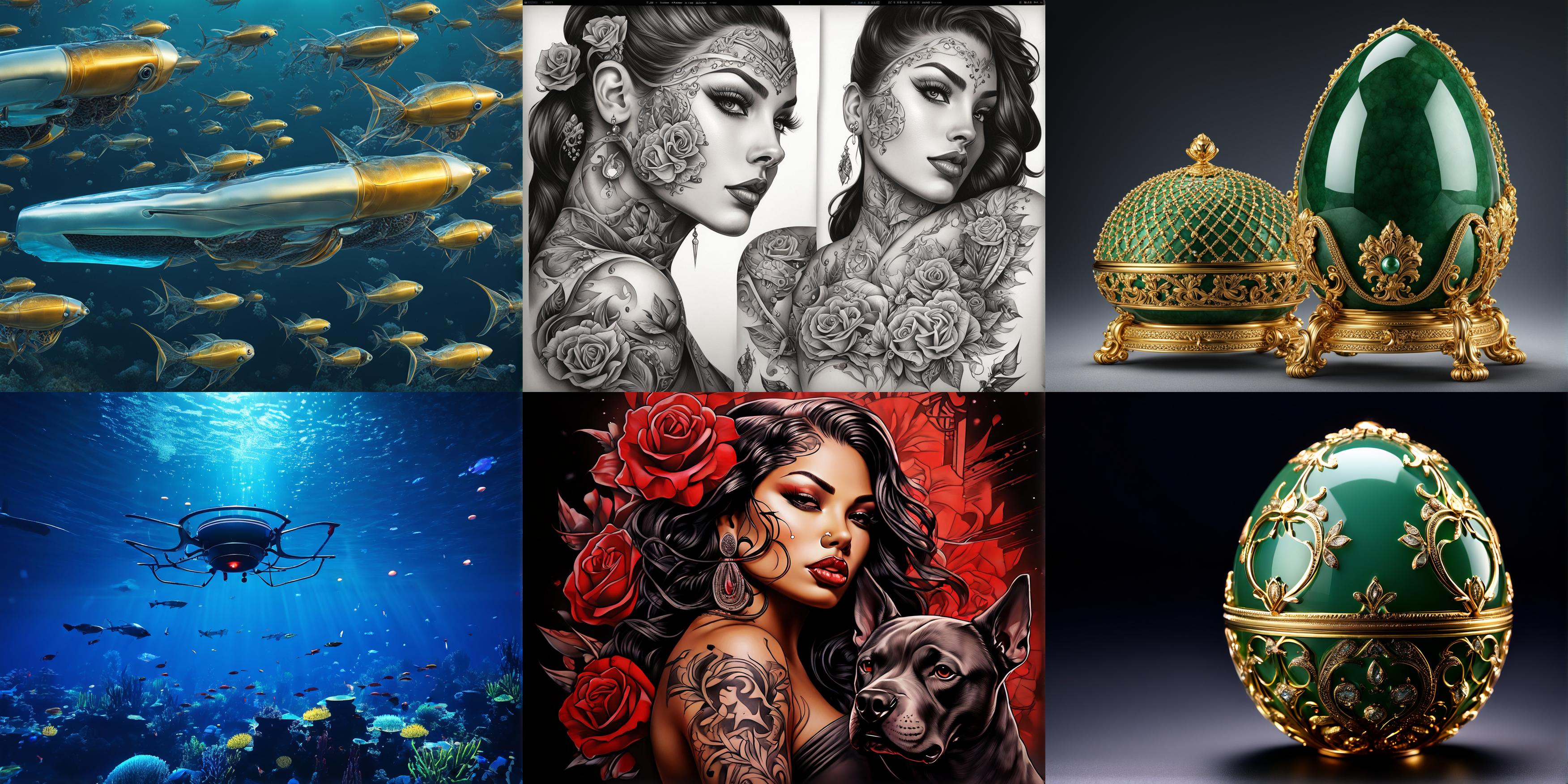}
\begin{tabular}{ccc}
    \multirow{6}{40mm}{\scriptsize{generate
an image of AI-controlled aquatic drones assisting marine biologists in studying the behavior of
plankton communities., Miki Asai Macro photography, close-up, hyper detailed, trending on artstation,
sharp focus, studio photo, intricate details, highly detailed, by greg rutkowsk.}} &  \multirow{2}{40mm}{\scriptsize{tattoo sketch chicano
style, beauty and pit bull, blood, dynamics.}} &  \multirow{8}{40mm}{\scriptsize{image of a jade green and gold coloured Fabergé
egg, 16k resolution, highly detailed, product photography, trending on artstation, sharp focus, studio
photo, intricate details, fairly dark background, perfect lighting, perfect composition, sharp features,
Miki Asai Macro photography, close-up, hyper detailed, trending on artstation, sharp focus, studio
photo, intricate details, highly detailed, by greg rutkowski."}} 
\end{tabular}

\vspace{28mm}

\caption{\textbf{Qualitative comparison of landscape aspect ratios.} Aspect ratio 4:3 ($1168\times876$), the top is SDXL, bottom is Playground v2.5. Our model can generate content following the prompt consistently under extreme aspect ratios, while SDXL sometimes fails to follow the prompt, or generates multiple objects under wide aspect ratios. 
}
\label{fig:multiar-horiz}
\end{figure*}
\subsection{Generation Across Multiple Aspect Ratios}
\label{multiar}

The ability to generate images of various different aspect ratios is an important feature in real-world applications of text-to-image models. However, common pre-training procedures \cite{rombach2022highresolution, podell2023sdxl} for these models start by training only on square images in the early stages, with random or center cropping. This technique is standard practice from conditional generative models trained on ImageNet \cite{dhariwal2021diffusion, nichol2021improved, karras2022elucidating}.

Theoretically, this should not pose a problem. A diffusion model like SDXL \cite{rombach2022highresolution, podell2023sdxl} consisting of mostly convolution layers – mimicking a Convolutional Neural Network (CNN) – should work with any input resolution at inference time, even if it was not trained on that particular resolution. This is due to the transition-invariant \cite{lecun1989generalization, he2015deep} property of CNNs. Unfortunately, in practice, diffusion models do not generalize well to other aspect ratios when only trained on square images, as pointed out by NovelAI \cite{novelai}.

To address this challenge, NovelAI proposes doing bucketed sampling, where images with similar aspect ratios are bucketed together in the same forward pass. SDXL \cite{podell2023sdxl} adopted this bucketing strategy and also added additional conditioning to specify the source and target image sizes.

SDXL’s conditioning strategy forced the model to learn to place the image’s subject at the center under different aspect ratios. However, due to an unbalanced distribution of the aspect ratio buckets \cite{podell2023sdxl} in SDXL’s dataset, i.e. the majority of the dataset’s images are square, SDXL also learned the bias of certain aspect ratios in its conditioning. 
Furthermore, images generated at non-square aspect ratios typically exhibit much lower quality than square images.

In Playground v2.5, one of our explicit goals was to make the model reliably produce high-quality images at multiple aspect ratios, as we had learned from the users that this is a critical element for a high-quality production-grade model. While we followed a bucketing strategy similar to SDXL’s, we carefully crafted the data pipeline to ensure a more balanced bucket sampling strategy across various aspect ratios. Our strategy avoids catastrophic forgetting and helps the model not be biased towards one ratio or another.

Fig~\ref{fig:multiar-oct} and Fig~\ref{fig:multiar-horiz} show a qualitative comparison between SDXL and Playground v2.5 across portrait and landscape aspect ratios, respectively. Our model can generate high-quality images under various aspect ratios without errors like generating multiple objects or wrong composition. 
\begin{figure*}
    
\includegraphics[width=1.0\textwidth]{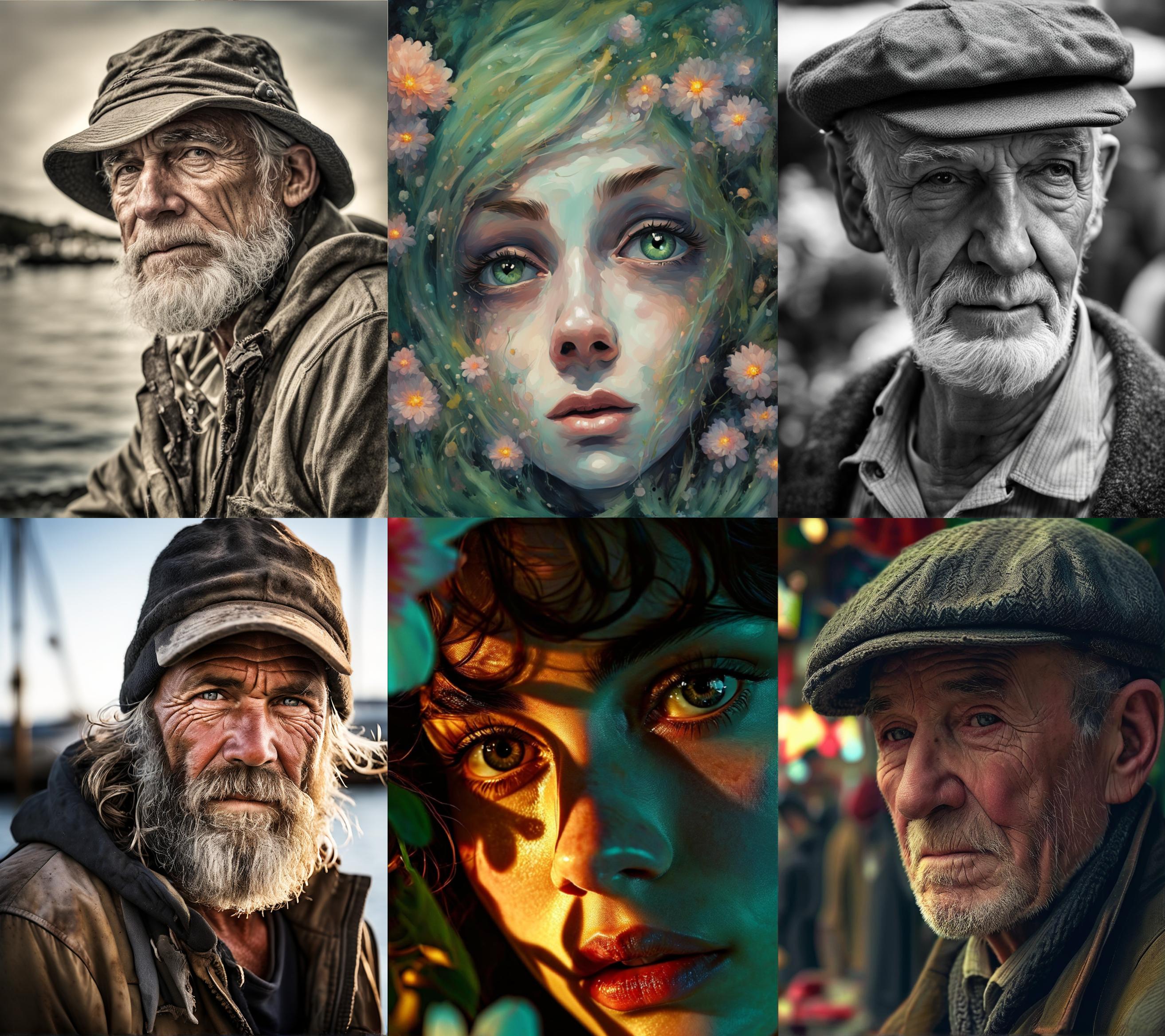}
\vspace{2mm}
{\setstretch{0.7}
\begin{tabular}{p{40mm}p{40mm}p{40mm}}
    {\scriptsize{Seasoned fisherman portrait, weathered skin etched with deep wrinkles, white beard, piercing gaze beneath a fisherman's hat, softly blurred dock background accentuating rugged features, captured under natural light, ultra-realistic, high dynamic range photo}} &
    {\scriptsize{Close-up portrait of a face with big eyes, overflowing with a mysterious, ethereal ambiance, capturing a direct gaze with the viewer amidst the chiaroscuro interplay of light and shadow, featuring soft colors alongside a splash of color from variously hued flowers, all set against a green retro Gothic night scene reminiscent of Mr X's style, face illuminated, reflective surfaces enhancing the rich, vivid textures, and the delicate details enhanced by octane rendering, cinematic portrayal}} &  
    {\scriptsize{An old man with a flat cap for his head in a market close up, trending on artstation, sharp focus, studio photo, intricate details, highly detailed, by greg rutkowski}} 
\end{tabular}
}

\caption{\textbf{Human aesthetic preference alignment comparison with SDXL.} Top row is SDXL, bottom row is Playground v2.5. Our model can generate better human-centric facial details, overall lighting, color and depth-of-field.
}

\label{fig:comp_details}

\end{figure*}
\begin{figure*}[t!]
\centering
\includegraphics[width=1.0\textwidth]{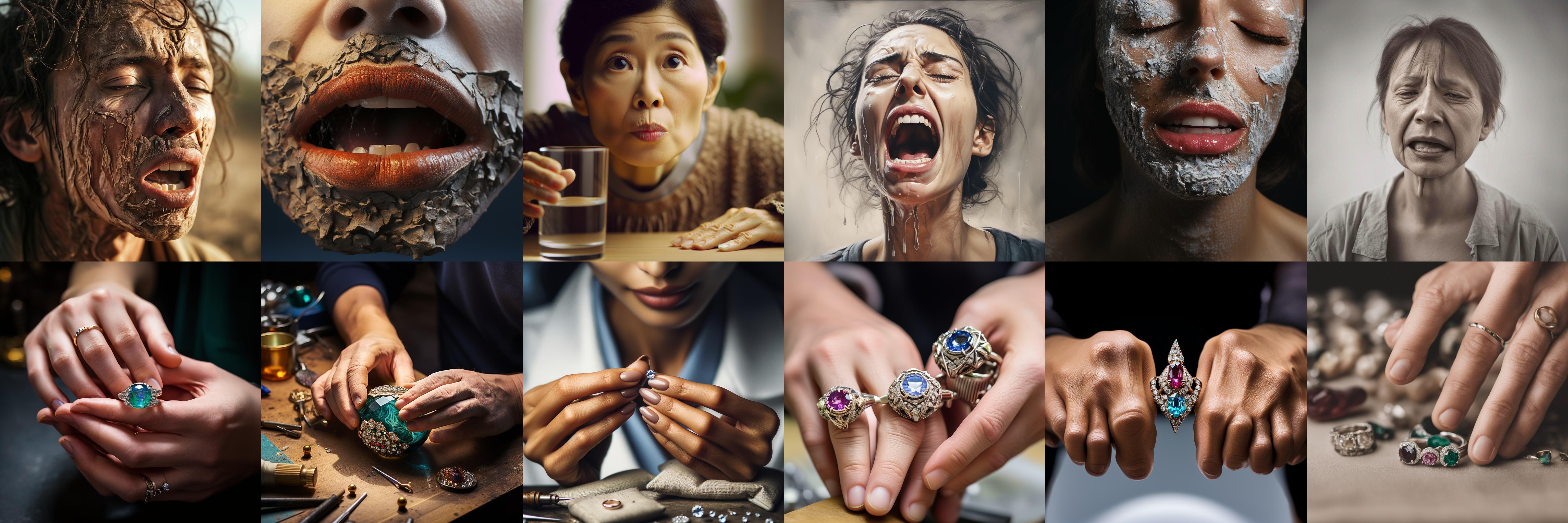}
\begin{tabular}{p{20mm}p{20mm}p{20mm}p{20mm}p{20mm}p{20mm}}
    \scriptsize{Playground v2.5} & \scriptsize{Midjourney 5.2} &  \scriptsize{DALL·E 3~\cite{betker2023improving}} & \scriptsize{Playground v2~\cite{playground-v2}} & \scriptsize{PixArt-alpha~\cite{chen2023pixart}} & \scriptsize{SDXL-1.0~\cite{podell2023sdxl}}
\end{tabular}
\caption{\textbf{Qualitative comparison between methods.} Prompts for the top row: "a person with a feeling of dryness in the mouth.", the bottom row: "a jeweler's hands, each holding a tiny gemstone, aligning them perfectly for a custom ring.". Our model can generate lively expressions, fine-details of a person's teeth, eyes, and expression, and the correct hands.
}
\label{fig:qual1}
\end{figure*}
\subsection{Human Preference Alignment} \label{human_pref_alignment}

Humans are particularly sensitive to visual errors on human features like hands, faces, and torsos. An image with perfect lighting, composition, and style will likely be voted as low-quality if the face of the human is malformed or the structure of the body is contorted.

Generative models, both language and image, are prone to hallucination. In image models, this can manifest as malformed human features. There are multiple reasons for hallucination, but one evident explanation is a misaligned training objective: generative models are trained to maximize the log-likelihood of the data rather than maximizing human preference. In LLMs, a common strategy to align pre-trained generative models with human preference is called supervised fine-tuning, or SFT. In short \cite{ouyang2022training}, SFT fine-tunes a pre-trained base model with a small but very-high-quality dataset. This simple technique often outperforms a more complicated approach like RLHF \cite{zhou2023lima}. However, the question of how to best curate an SFT alignment dataset from different sources to maximize performance on a downstream task remains an ongoing research problem \cite{xia2024less}.

One of our goals with Playground v2.5 was to reduce the likelihood of visual errors in human features, which is a common critique of open-source diffusion models more broadly, as compared to closed-source models like Midjourney. Emu \cite{dai2023emu} introduces an alignment strategy similar to SFT for text-to-image generative models. Inspired by Emu, we developed a system that enables us to automatically curate a high-quality dataset from multiple sources via user ratings. Then, we took an iterative, human-in-the-loop \cite{parkash2012attributes,Mahmood_2022_CVPR} training approach to select the best dataset candidates. We monitored the training progress by empirical evaluation, comparing our aligned models by looking at image grids generated from a fixed set of prompts, similar to \cite{zhou2023lima}.


Our novel alignment strategy enables us to excel over SDXL in at least four important human-centric categories:

\begin{itemize}[noitemsep]
  \item Facial detail, clarity, and liveliness
  \item Eye shape and gaze
  \item Hair texture
  \item Overall lighting, color, saturation, and depth-of-field
\end{itemize}

We chose to focus on these categories based on usage patterns in our product and from user feedback.

In Fig~\ref{fig:comp_details}, we showcase some examples of the difference in fine details between images generated using Playground v2.5 and those using SDXL. In Fig~\ref{fig:qual1}, we compare our model with other SoTA methods in generating human-centric images.

\begin{figure}[t!]
\centering
\includegraphics[width=0.8\textwidth]{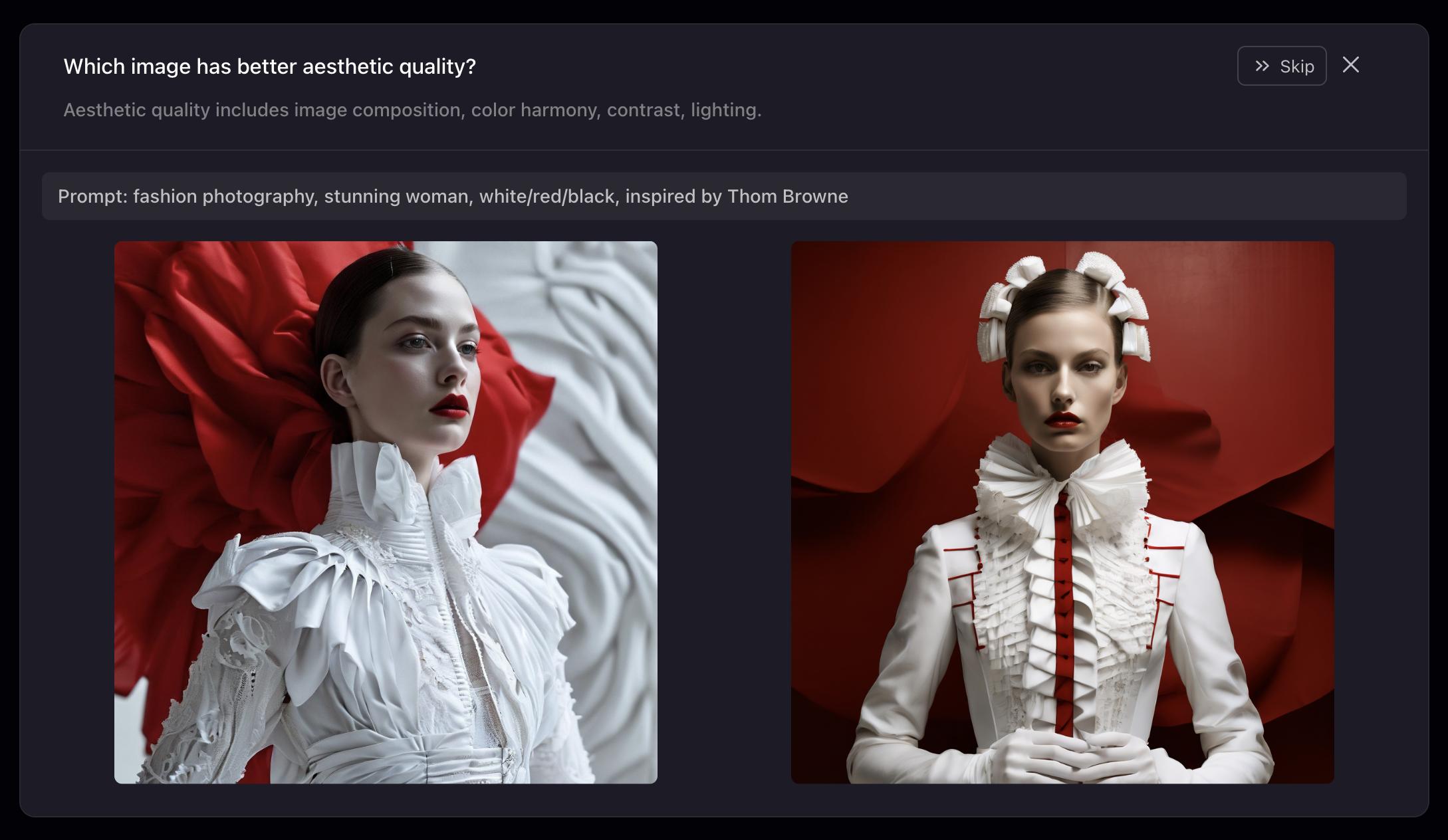}
\caption{An example of an image pair shown to a user in our product.}
\label{fig:user_study}
\end{figure}

\section{Evaluations}
\label{eval}
\subsection{User Study Interface}

Since we ultimately build our models to be used by our hundreds of thousands of users, it is critical for us to understand their preferences on model outputs. To this end, we conduct our user studies directly within our product (see Fig \ref{fig:user_study}). We believe this is the best context to gather preference metrics and provides the harshest test of whether a model actually delivers on making something valuable for an end user.	

For a given user study, we choose a fixed set of prompts and sample images from two models. We then show a pair of images with the prompt to the user (without showing them which model corresponds to which image) and ask them to pick the best one according to some attribute, e.g. aesthetic preference. Because a single user’s rating is prone to bias, we show each image pair to at least 7 unique users. To further reduce bias, an image pair only "wins" for a given model if its output is preferred by at least a 2-vote margin. A 1-vote margin is considered a tie. Lastly, we involve thousands of unique users on each user study. All user studies mentioned in this report are conducted through this interface.

We conducted studies to measure overall aesthetic preference, as well as for the specific areas we aimed to improve with Playground v2.5, namely generation across multiple aspect ratios and human preference alignment. 

\subsection{Overall Aesthetic Preference against other SoTA models}
\label{sota_study}
\begin{figure*}[t!]
\centering
\includegraphics[width=1.0\textwidth]{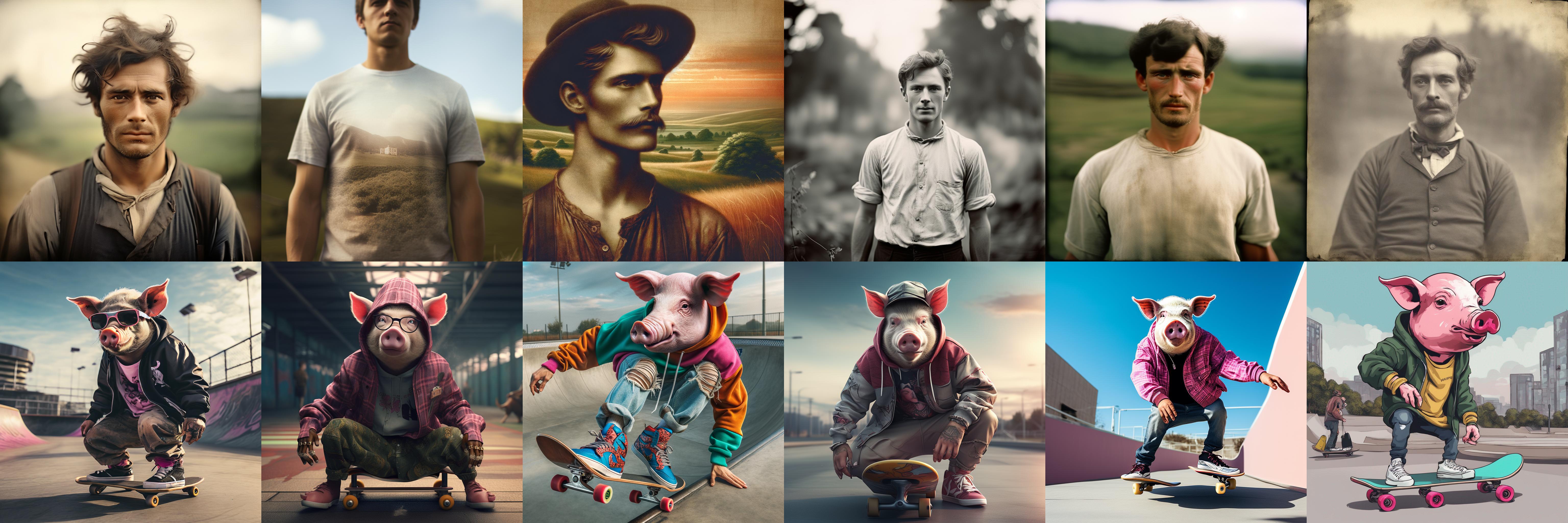}
\begin{tabular}{p{20mm}p{20mm}p{20mm}p{20mm}p{20mm}p{20mm}}
    \scriptsize{Playground v2.5} & \scriptsize{Midjourney 5.2} &  \scriptsize{DALL·E 3~\cite{betker2023improving}} & \scriptsize{Playground v2~\cite{playground-v2}} & \scriptsize{PixArt-alpha~\cite{chen2023pixart}} & \scriptsize{SDXL-1.0~\cite{podell2023sdxl}}
\end{tabular}
\caption{\textbf{More qualitative comparison between methods.} Prompts for the top row: "blurred landscape, close-up photo of man, 1800s, dressed in t-shirt", the bottom row: "human with pig head wearing streetwear fashion clothes, skateboarding on a skatepark". 
}
\label{fig:qual2}
\end{figure*}
\begin{figure*}[t]
\centering
\vspace{-5mm}
\includegraphics[width=0.95\textwidth]{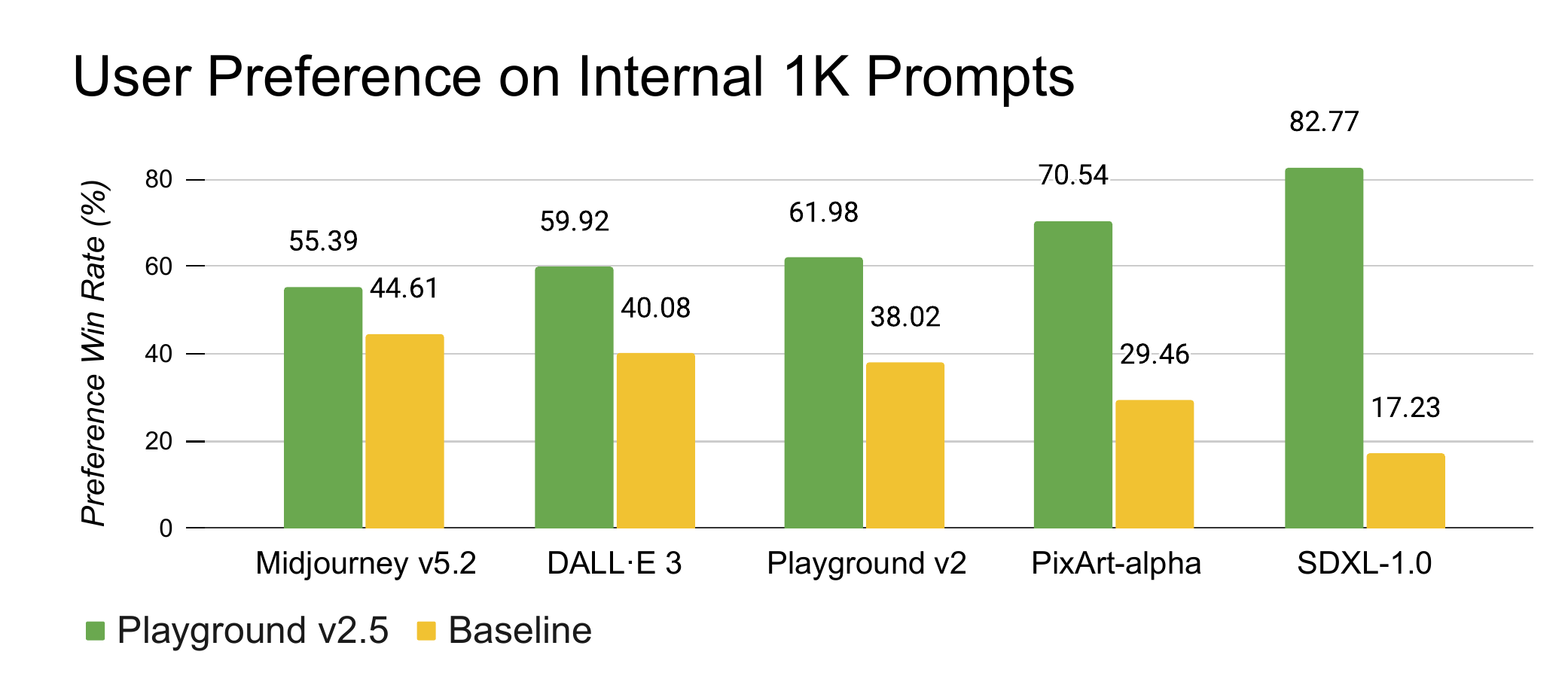}
\caption{\textbf{User study against SoTA Methods.} We report human aesthetic preference metrics of Playground v2.5 against various publicly available text-to-image models. Playground v2.5 outperforms Midjourney 5.2, DALL{$\cdot$}E 3 \cite{betker2023improving}, Playground v2 \cite{playground-v2}, PIXART-$\alpha$ \cite{chen2023pixart}, and SDXL \cite{podell2023sdxl}.
}
\label{fig:sota}
\end{figure*}

We used a prompt set called Internal-1K to compare Playground v2.5 against other state-of-the-art models. Internal-1K is a prompt set collected from real user prompts on Playground.com, making it representative of real users’ prompting styles. We showed image pairs to thousands of users, specifically focusing on aesthetic preference for this study. This is the same study setup as our previous release of Playground v2 \cite{playground-v2}. For reference, our previous studies demonstrated that images produced from Playground v2 were favored 2.5x more than those produced by SDXL. We aimed to surpass this for Playground v2.5 and succeeded: v2.5 is favored 4.8x over SDXL.

Fig. \ref{fig:sota} shows our results against various publicly available text-to-image models. Across the board, Playground v2.5 dramatically outperforms the current state-of-the-art open source models of SDXL~\cite{podell2023sdxl} and PIXART-$\alpha$~\cite{chen2023pixart}, as well as Playground v2~\cite{playground-v2}. Because the performance differential between Playground v2.5 and SDXL was so large, we also tested against state-of-the-art closed-source models like DALL$\cdot$E 3~\cite{betker2023improving} and Midjourney 5.2, and found that Playground v2.5 still outperforms these models in aesthetic quality.

\begin{figure*}[t]
\centering
\vspace{-5mm}
\includegraphics[width=0.95\textwidth]{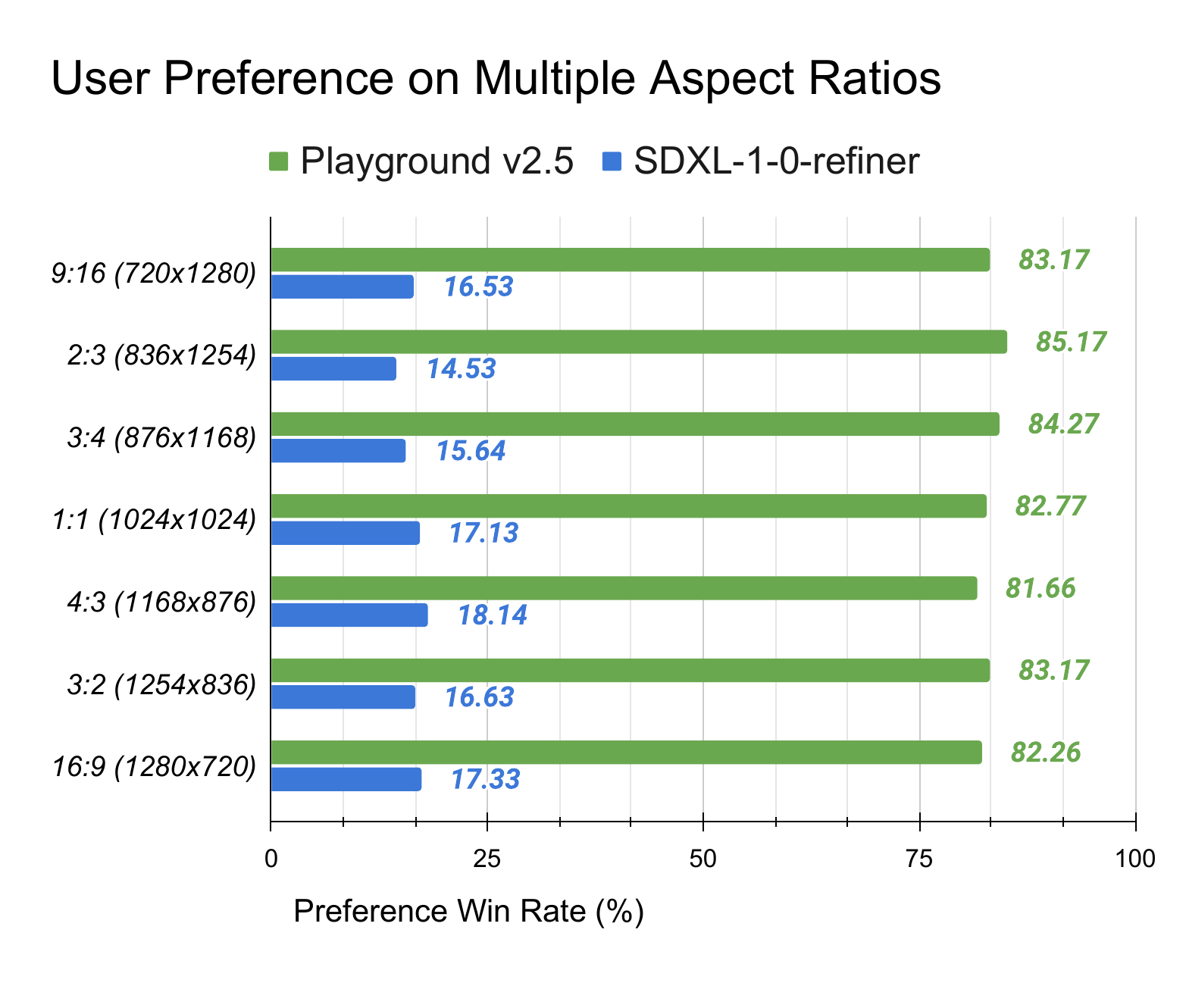}
\vspace{-5mm}
\caption{\textbf{User study against SDXL on multiple aspect ratios.} We conduct user studies for images generated in various commonly-used aspect ratios, height-to-width ratios ranging from 9:16 to 16:9. Our model outperforms SDXL in all aspect ratios by a large margin.
}
\label{fig:multiar}
\end{figure*}

\subsection{Evaluation of Generation Across Multiple Aspect Ratios}

We report user preference study metrics on commonly-used aspect ratios using the Internal-1K prompt set. We conducted a separate user study for each aspect ratio, ranging from 9:16 to 16:9. For a given study, we used the same aspect ratio conditioning for both models on all images. Fig \ref{fig:multiar} shows our results. Our model outperforms SDXL in all aspect ratios by a large margin.

\subsection{Evaluation on People-centric Prompts}

\begin{figure*}[t]
\centering
\vspace{-5mm}
\includegraphics[width=0.8\textwidth]{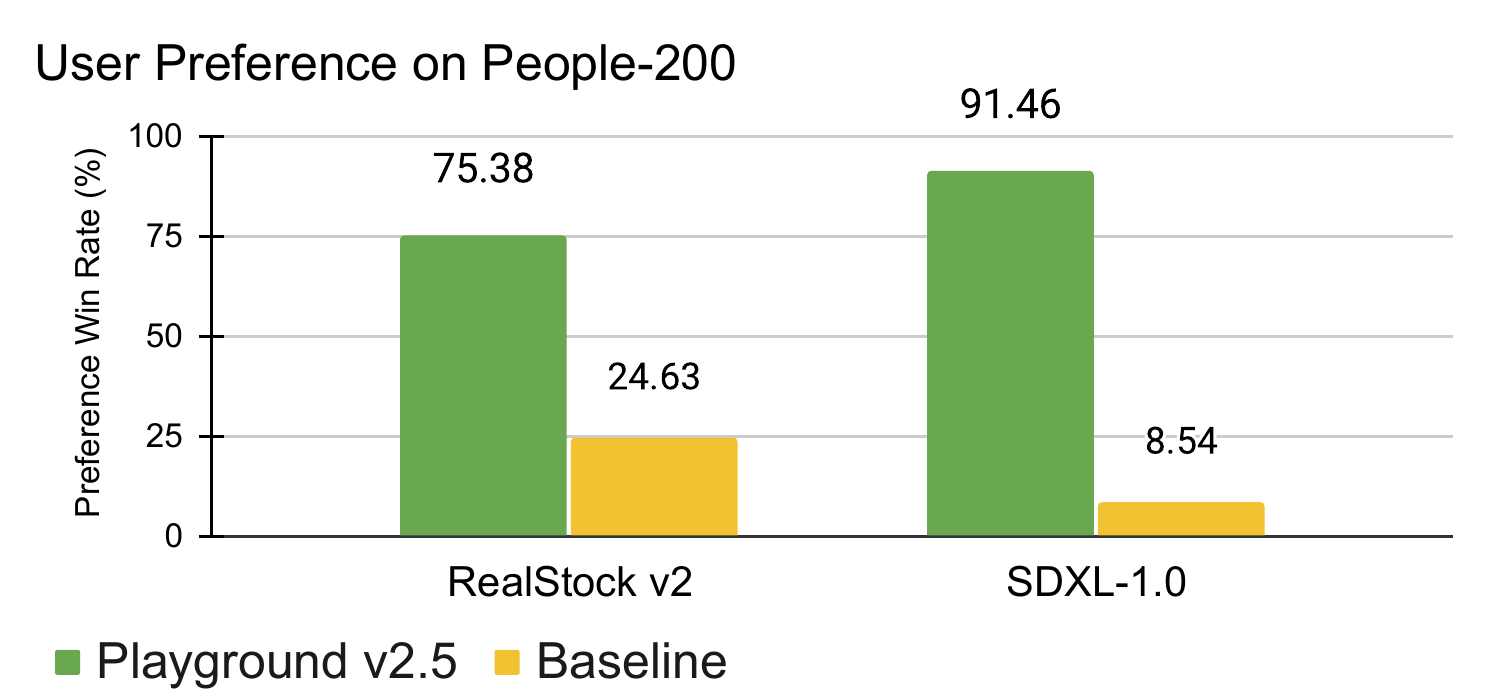}
\caption{\textbf{People-200 benchmark.} We conduct a user study using the People-200 prompt set, which focuses on generating people. We compare Playground v2 against two baseline models: SDXL\cite{podell2023sdxl} and RealStock v2, a popular community fine-tune trained on a realistic people dataset. All images were generated in 3:2 aspect ratios with resolution 1254x836.
}
\label{fig:people200}
\end{figure*}

As discussed in Section \ref{human_pref_alignment} about improving human preference alignment, people-related prompts are an important practical use-case for commercial text-to-image models. Indeed, they are quite popular in our product. To assess our model’s ability to generate people-related images, we curated 200 high-quality people-related prompts from real user prompts in our product. We call this the \textit{People-200} prompt set. We will release this prompt set to the community for benchmarking purposes.

We conducted our user study using portrait aspect ratio 3:2, since this is the most popular choice in the community for images showing people. We compared Playground v2.5 against two commonly-used baseline models: SDXL and RealStock v2, a community fine-tune of SDXL that was trained on a realistic people dataset.

Fig \ref{fig:people200} shows that Playground v2.5 outperforms both baselines by a large margin.

\subsection{Automatic Evaluation Benchmark}
\label{mjhq30k}

\begin{figure*}[t]
\centering
\vspace{-5mm}
\includegraphics[width=1.0\textwidth]{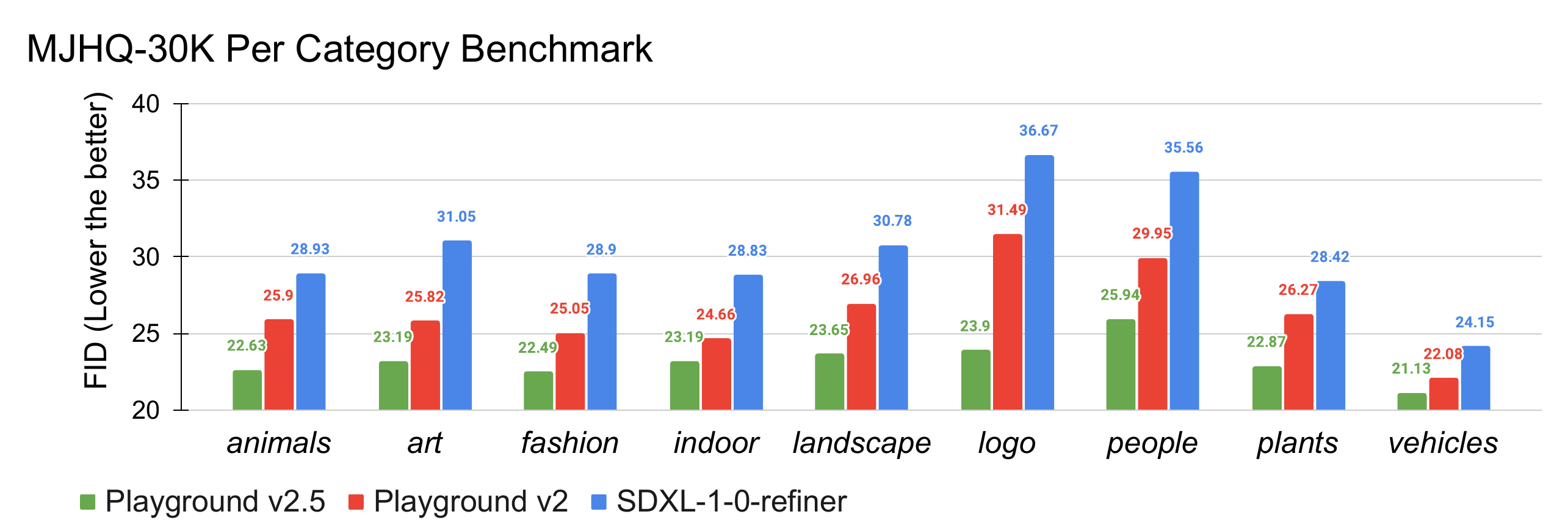}
\caption{\textbf{MJHQ-30K benchmark.} We report FID of Playground v2.5, Playground v2 \cite{playground-v2}, and SDXL\cite{podell2023sdxl} on 10 common categories. Playground v2.5 outperforms Playground v2 in all categories, and most significantly on \textit{logo} and \textit{people} categories.
}
\label{fig:mjhq30k}
\end{figure*}

Lastly, we introduce a new benchmark, \textit{MJHQ-30K}, for automatic evaluation of a model’s aesthetic quality. The benchmark computes Fréchet Inception Distance (FID) \cite{heusel2018gans} on a high-quality dataset to gauge aesthetic quality. By spot-checking the FID and ensuring it was trending lower, we were able to quickly gauge progress throughout the different stages of pre-training and alignment.

We curated a high-quality dataset from images made on Midjourney 5.2. The dataset covers 10 common categories, and each category has 3K samples. Following common practice, we used aesthetic score \cite{kirstain2023pickapic} and CLIP score \cite{hessel2022clipscore} to ensure high image quality and high text-to-image alignment. 
\begin{wraptable}{8r}{0.45\textwidth}
\vspace{-2mm}
\centering
\setlength{\tabcolsep}{6pt}
\renewcommand{\arraystretch}{1.2}
\caption{\textbf{MJHQ-30K overall FID.}}
\begin{tabular}{c|c}
\hline
    Method & Overall FID\\ \hline
    SDXL 1.0 + refiner\cite{podell2023sdxl} & 9.55 \\ \hline
    Playground v2 \cite{playground-v2} & 7.07 \\ \hline
    Playground v2.5 & \textbf{4.48} \\ \hline
\end{tabular}
\label{tbl:mjhq30k}
\end{wraptable}
Furthermore, we took extra care to make the images and prompts well-varied within each category.

We report both the overall FID (Table \ref{tbl:mjhq30k}) and per category FID (Fig \ref{fig:mjhq30k}). All FID metrics are computed at resolution 1024x1024. Our results show that Playground v2.5 outperforms both Playground v2 and SDXL in overall FID and all category FIDs, especially in the people and fashion categories. This is in line with the results of the user study, which indicates a correlation between human preferences and the FID score of the MJHQ30K benchmark.

We release this benchmark to the public on HuggingFace\footnote{ \url{https://huggingface.co/datasets/playgroundai/MJHQ-30K}} and encourage the community to adopt it for benchmarking their models’ aesthetic quality during pre-training and alignment.

\begin{figure*}[b]
\centering
\vspace{-5mm}
\includegraphics[width=1.0\textwidth]{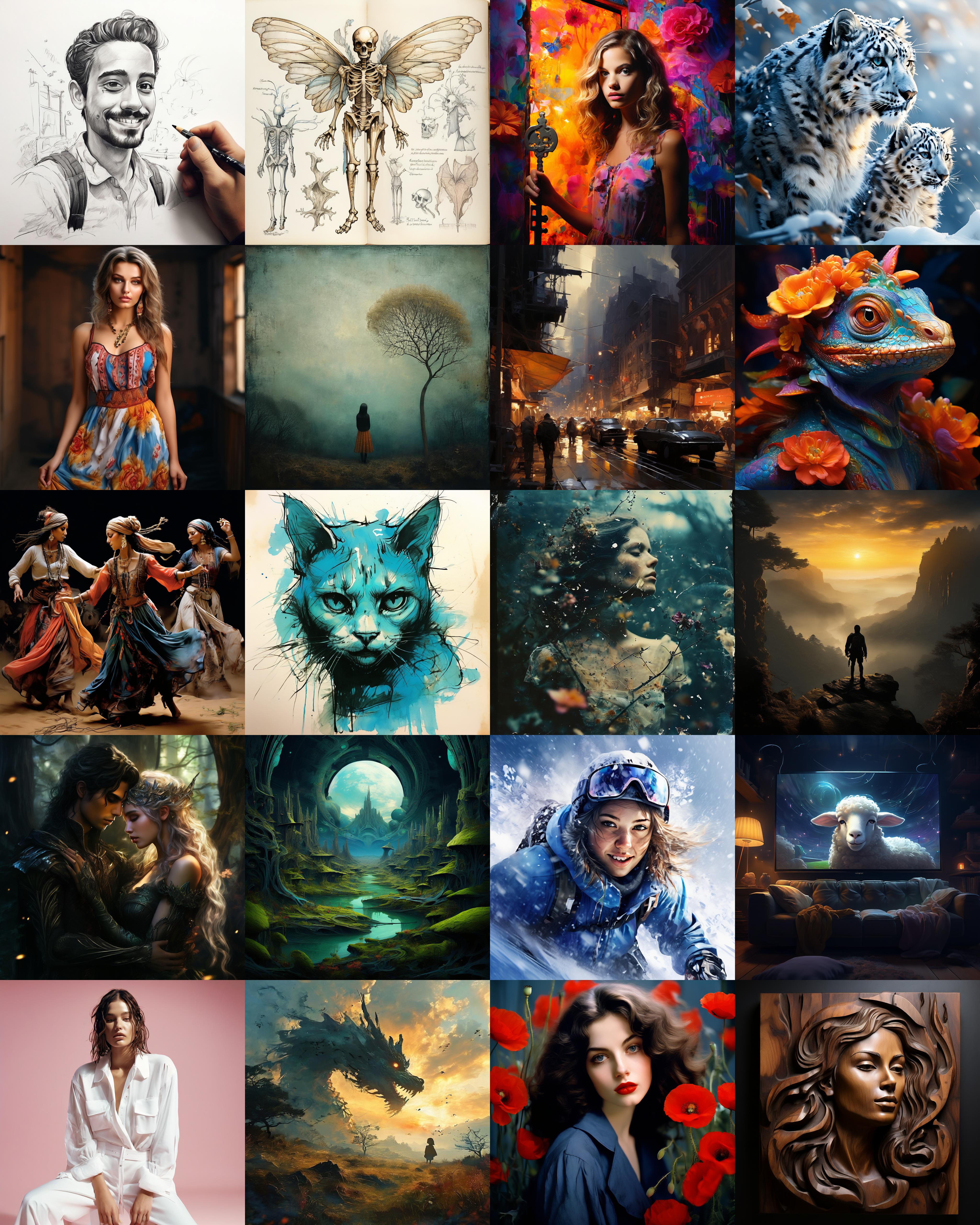}
\caption{Playground v2.5 random samples with popular user prompts.}

\label{fig:pgv25_random}
\end{figure*}





\section{Conclusion}
In this work, we share three insights for achieving state-of-the-art aesthetic quality
in text-to-image generative models, and we analyze and empirically evaluate Playground v2.5 against SoTA models in various conditions and setups. Playground v2.5 demonstrates: (1) superior performance in enhancing image color and contrast, (2) ability to generate high-quality images under various aspect ratios, and (3) alignment to human preference for aesthetic quality in generated images, especially for fine details in images of humans.

We are excited to release Playground v2.5 to the public. The model is available today to use at our product website\footnote{\url{https://playground.com}} for all users, and we have open-sourced the weights on HuggingFace\footnote{\url{https://huggingface.co/playgroundai/playground-v2.5-1024px-aesthetic}}. Furthermore, we will soon provide extensions for using Playground v2.5 in A1111 and ComfyUI, two popular community tools for using diffusion models.

For future works, we hope to tackle improving text-to-image alignment, enhancing the model's variation capabilities, and exploring new architectures. 


At Playground, our goal is to build a unified general-purpose vision system that deeply understands pixels and enables humans of all skill levels to masterfully generate and edit pixels. We see Playground v2.5 as a stepping stone towards this vision, and we encourage the community to build with us.

\clearpage
{\small
\bibliographystyle{plain}
\bibliography{egbib}
}


\end{document}